\documentclass[10pt,letterpaper]{article}

\usepackage{cogsci}

\cogscifinalcopy % Uncomment this line for the final submission 

\usepackage{pslatex}
\usepackage{apacite}
\usepackage{float} % Roger Levy added this and changed figure/table
\usepackage{tcolorbox}
\usepackage{color, colortbl}
\usepackage{multirow}
\usepackage{booktabs}
\usepackage{amsmath}
\usepackage{tabularx}
\usepackage{makecell}
\title{On Reality and the Limits of Language Data:\\ Aligning LLMs with Human Norms}

\author{{\large \bf Nigel H. Collier$^1$ \ \ \ \  Fangyu Liu$^1$ \ \ \ \ Ehsan Shareghi$^2$} \\
$^1$University of Cambridge, UK \ \ \ \ \ $^2$Monash University, Australia \\
 \ \ \ \ \texttt{\{nhc30,fl399\}@cam.ac.uk} \ \ \ \  \texttt{ehsan.shareghi@monash.edu}
}

\begin{document}
\maketitle
\begin{abstract}
Recent advancements in Large Language Models (LLMs) harness linguistic associations in vast natural language data for practical applications. However, their ability to understand the physical world using only language data remains a question.  After reviewing existing protocols, we explore this question using a novel and tightly controlled reasoning test (ART) and compare human norms against versions of GPT-3. Our findings highlight the categories of common-sense relations models that could learn directly from data and areas of weakness.  GPT-3 offers evidence for verbal reasoning on a par with human subjects for several relations including Synonymy, Antonymy, and Default inheritance, Without reinforcement learning from human judgements, it appears GPT-3 performs at the lower end of the reference interval for Has-part and Contained-in. Weaknesses were observed also in affordance characteristics through Necessary-quality, Order-of-size and Order-of-intensity. Combining LLMs with symbolic world grounding is a promising direction to address associative learning.

\end{abstract}

\section{Introduction}

We consider 
the relationship between the real world and the meaning representations that are
derived from large-scale language data using large language models (LLMs) that have been trained using self-supervision, i.e. learning to predict a part of the input from the other parts. 

A widely discussed problem in the Philosophy of Language concerns the power of language to represent reality - the world of \lq sticks and stones; cats and trees' \cite{devitt1999language}, of \lq molecules, galaxies, and babies' \cite{searle2010basic} that we know about from advances in scientific discovery. This question  
finds voice in 
recent works that try to close the gap between LLMs and normative human performance on language tasks~\cite{wang2019superglue,srivastava2022beyond}.

Do the limitations of LLMs arise from data or fundamental properties inherent to their structure? A variety of evaluation metrics~\cite{gatt2018survey} across multiple tasks and models have been pursued to gain insight into this question \cite{tenney2019bert,rogers2020primer}. Whilst evaluations have spurred remarkable achievements in many tasks, we argue that they obscure a fundamental question about the limitation of LLMs to represent reality. We
look at these models' capacity to tell true from false propositions about \emph{na\"ive physics}, a subset of what is called common sense  reasoning~\cite{da2019cracking,rae2021scaling,lin2021truthfulqa}, that focuses on the intuitive truth that human cognition endorses about the physical world \cite{smith1994naive}. 

Whilst common sense reasoning has long been considered a part of the natural language understanding evaluations, e.g. the Winograd Schema Challenge (WSC) \cite{levesque2012winograd}, the exact kinds of knowledge that model builders should target is rarely made explicit. 
 
The characteristic of reasoning with \emph{expressed} facts along with \emph{unstated} facts (enthymemes) from a store of common sense knowledge appears fundamental to human understanding~\cite{singer1992validation}. Likewise, LLMs will require access to enthymematic mechanisms if they are to make predictions about and correctly {\it explain} the long tail of \emph{unstated} facts not explicitly mentioned in training data. We triangulate between three synergistic fields: philosophical ontology, cognitive psychology and NLP. A philosophical approach to objective reality is taken to motivate some of the necessary goals of common sense reasoning.  We then present a psychologically motivated
test, based on analogical reasoning  to interpret model capabilities. The analogy test focuses on understanding word meanings when presented as ordered pairs, i.e. of the form {\it a is to b as c is to d}. Analogical Reasoning Tests (ARTs) such as the Miller Analogies Test \cite{house1996differential} are widely trusted as indicators of human academic achievement and have long been viewed in AI as crucial instruments to elucidate the role of knowledge in problem solving \cite{minsky1975framework,lenat1985cyc,hall1989computational}. 

We present a new bipartite ART test, ART A\&B,\footnote{Available to download in BigBench json format from
\url{https://sites.google.com/site/nhcollier/projects/art}
} consisting both of analogy questions in Section A
and a structured set of semantic relation questions in Section B 
(covering 16 semantic relations) that underlie the analogy test.  
The test is intended for use in zero-shot settings to focus on true generalization from foundation knowledge.
Additionally, we use the test responses of 61 participants\footnote{Native English speaker residents of the UK.} to construct two normative profiles: the first against human expert judgements and the second against a common sense judgement where at least 80\% of participants agreed. We report the results of 
five GPT-3 language models and ChatGPT against the two profiles. To rule out results based on associative behaviour we further examined the results of nonsymmetric semantic relation items where the largest models (text-Davinci-002, code-Davinci-002 and ChatGPT) fell within human norms. 
We observe that the three smallest GPT-3 models fall outside both expert and common sense norms. After ruling out associative behaviour, the text-Davinci-002 falls within norms for half the relations but at the lower end or outside human norms for relations that require understanding of mereotopological and affordance relations.  ChatGPT which is augmented with data selection and reinforcement learning, 
performs far better at understanding object qualities but still falls at the lower end of human performance for several mereotopological relations.

We hope that this test will help to clarify several misconceptions about common sense reasoning in current state-of-the-art LLMs, draw attention to algorithmic deficiencies, and facilitate coherent strategies for achievement outcomes. 

\section{Related work}

Here we briefly review the case for why na\"ive physics is a core part of common sense, why corpora might be deficient in allowing us to reconstruct common sense, and provide a survey of important common sense benchmarks. 

\subsubsection{What is common sense?} 

Despite its apparently obvious nature, common sense remains an uneasy obscurity in the NLP literature which often describes it via its absence, e.g. as the output of a system that elicits assertions within a large corpus~\cite{schubert2002can} or as a database \cite{angeli2014naturalli}. 
Arguably, a critical examination of {\it what is common sense knowledge?} and its distinction from other types of knowledge feels long overdue if we are to construct productive research questions, e.g. how stable is such knowledge? how malleable are LLMs for representing such knowledge? A satisfying discussion is beyond the scope of this work but for now we highlight Brachman and Levesque's \shortcite{brachman2021toward} working definition: {\it Common sense is the ability to make effective use of ordinary, everyday, experiential knowledge in achieving ordinary, practical goals.} 
The definition, with parallels to Gibson's world of affordances \shortcite{gibson1979ecological}, is ambitious and draws attention to the need for models to act holistically upon common sense knowledge within everyday tasks. We favour this as an end goal but within this complex endeavour~\cite{davis2015commonsense} we are concerned to systematically investigate the essential core of non-holistic common sense knowledge, i.e. commonly held knowledge about physical reality which is capable of scientific treatment. Recent commentaries on human intelligence support this view that an intuitive understanding of the physical world is core in human infants and animals but currently lacking in artificial intelligence~\cite{hassabis2017neuroscience,crosby2020animal}. 

In ontological terms what might {\it obvious} physical facts be? It is a difficult question but in the tradition of philosophical realism, Smith \shortcite{smith1995formal,milton2004top} has argued for a stable and culturally invariant notion of common sense. Following this, the common sense we seek to capture is a subset of objective reality - independent of an individual's cognition - to which ordinary~(non-expert) cognition can relate via contact with the physical environment we perceive.

\noindent{\bf Are corpora Gestalt sources of common sense knowledge?}   A 
premise in NLP is that LLMs might be able to recover 
implicit physical common sense knowledge \cite{forbes2019neural}. However, when thinking about facts reported in large-scale corpus data, as Russell noted \cite{russell1914relation}, what might be commented on is not the whole sense at one time but rather the part that individuals single out for attention. 

When we come to analyse factual descriptions arising from perceptual evidence within a population it seems natural that we should expect variation due to individual's thinking disposition.  A question therefore arises: when the pieces of perceptual evidence are combined from a large group of individuals we find on the Web, does a combination of data, the algorithm and the physical computer \cite{kaplan2020scaling} allow an LLM to represent the sum of common sense reality as precisely as humans perceive it
from the environment? 
Such an important question is almost never explicitly posed in
probes of natural language generation (NLG) which 
often lack sufficiency for semantic proposition content~\cite{anderson2016spice}.

\noindent{\bf Common Sense Benchmarks.} A substantial body of literature exists that observes how models make predictions on common sense benchmarks. For a recent and comprehensive survey we refer readers to Storks {\it et al.} \shortcite{storks2019recent}. One of the most influential evaluations is the Winograd Schema Challenge (WSC) \cite{levesque2012winograd}, a discriminative alternative to the generative probe set out in Turing's 1950's Imitation Game \shortcite{turing2009computing}. Each question in the test has an ambiguous pronoun and a binary choice of antecedents, for example, {\it The trophy doesn't fit in the brown suitcase because {\bf it's} too big. Options: {\bf trophy} or suitcase.} The example is designed so that humans easily resolve the ambiguity based on their understanding that part of the physical and social reality of {\it suitcase} is its use as a container. 

The WSC has inspired much work, however it has been criticised on a variety of grounds \cite{kocijan2020review} such as the high cognitive load required to craft the examples, the limited range of test sentences, the sensitivity of some examples to selectional restrictions, the presence of switchable referents, and the possibility to employ simple associative tricks \cite{kocijan2019surprisingly} based on a high mutual information between a trigger verb and an antecedent. Despite recent applications of adversarial filtration to create Winogrande debiased (WG) a large-scale version of the WSC that is immune to associative solutions \cite{sakaguchi2020winogrande}, the limitations of WSC call into question the external validity of the test and whether system performance on the WSC is indicative of advances in common sense reasoning \cite{trichelair2019reasonable}.  In response to these limitations, 
alternative benchmarks have emerged which vary greatly in style and complexity such as common sense question answering with OpenBook QA \cite[OBQA]{mihaylov2018can}, CommonsenseQA \cite[CSQA]{talmor2019commonsenseqa}, Cosmos QA \cite[CQA]{huang2019cosmos}, and Commonsense QA 2.0 \cite[CSQA2]{talmor2022commonsenseqa}, everyday situation understanding \cite[SWAG]{zellers2018swag}, and likely causes of sentence endings with the Choice of Plausible Alternatives \cite[COPA]{roemmele2011choice}. Closest to our work is PROST \cite{aroca2021prost}, 
%a recent benchmark 
focusing on physical common sense which uses an alternative task construction to the one we present here: a small closed world of objects serves as the source for the generation of multiple choice questions using a set of fixed templates. Ten relations explore model understanding of object attributes and affordances. As reported in Table~\ref{tab:datasets} (upper part), 
whilst previous benchmarks have provided insights into common sense reasoning by LLMs, these insights are often limited to the data set due to their holistic nature and low level of ontological control. 

\begin{table}[t!]
    \centering\small
        \caption{Attributes of modern common sense reasoning benchmarks. $N$ denotes the relation did not appear to be explicitly controlled and vice versa. $^{1}$WordNet is included for reference; $^{2}$Social common sense corresponds to linguistic/interpersonal intelligence \cite{humphrey2007society} and Physical common sense corresponds to logical-mathematical/naturalistic intelligence; $^{3}$Expert selection of objects and question patterns followed by automated generation; $^{4}12.2k$ Obtained after de-biasing; $^{5}$Hardening the data  through bias reduction; $^{6}$Hardening the data through  distractor answers; $^{7}$Hardening  the data through gamification; $^{8}$Hardening of the data set is done indirectly via restriction to zero-shot setting to avoid biasing training to testing data.} 
    \label{tab:datasets}
\scalebox{0.65}{
\begin{tabular}{l p{0.7cm} p{0.7cm} p{0.7cm} p{0.9cm} p{0.8cm} p{0.8cm} p{0.9cm} p{0.7cm}}
\Xhline{4\arrayrulewidth}
   & {\bf WSC} & {\bf WG} & {\bf CSQA} & {\bf CSQA2} & {\bf COPA} & {\bf WNet$^{1}$} & {\bf PROST} & {\bf $\text{ART}_{\text{A\&B}}$}\\
 %     & &  & & & &  &   & {A\&B}\\
\midrule
Physical$^{2}$ & Y & Y & Y & Y & Y & Y & Y & Y \\
Social$^{2}$ & Y & Y & Y & N & Y & Y & N & N \\
\midrule
Source & Expert & Crowd& Crowd& Crowd & Expert& Expert & Expert$^{3}$ & Expert \\
\midrule
Size & 273 & 44k$^{4}$ & 9.5k & 14.3k & 1K & 207k & 18.7k & 368 \\
\midrule
Hardening & N & Y$^{5}$ & Y$^{6}$ & Y$^{7}$ & N & N & Y$^{8}$ & Y$^{8}$ \\
\midrule
\multicolumn{9}{c}{{\bf Explicit relation controls}} \\
Synonym & N & N & N & N & N & Y & N & Y \\
Necessary quality & N & N & N & Y & N & N & N & Y \\
Associated quality & N & N & N & Y & N & N & N & Y \\
Has part & N & N & Y & Y & N & Y & N & Y \\
Order of intensity & N & N & N & N & N & N & Y & Y \\
Order of size & N & N & N & Y & N & N & Y & Y \\
Cause and effect & N & N & Y & Y & Y & N & N & Y \\
Capable of & N & N & Y & Y & N & N & Y & Y \\
Default inheritance & N & N & N & N & N & N & N & Y \\
Precondition & N & N & Y & Y & N & N & N & Y \\
Boundary (bona fide) & N & N & N & N & N & N & N & Y \\
Antonym & N & N & Y & N & N & Y & N & Y \\
Contained in & N & N & N & N & N & N & N & Y \\
Has member & N & N & Y & N & N & Y & N & Y \\
Troponym & N & N & N & N & N & Y & N & Y\\
Boundary (fiat) & N & N & N & N & N & N & N & Y \\
\Xhline{4\arrayrulewidth}
        \multicolumn{9}{c}{{\bf Example prompts for each relation}} \\
Synonym:& \multicolumn{8}{l}{{\it Do congeal and thicken have a similar meaning?}}\\
Necessary quality: &\multicolumn{8}{l}{{\it Must a rope be bendy?}}\\
Associated quality:&\multicolumn{8}{l}{{\it Can a mug be mined?}}\\
Has part:          &\multicolumn{8}{l}{{\it Is a core a part of an apple?}}\\
Order of intensity:&\multicolumn{8}{l}{{\it Is coal shinier than aluminium foil?}}\\
Order of size:&\multicolumn{8}{l}{{\it Is a country longer than a town?}}\\
Cause and effect: &\multicolumn{8}{l}{{\it Will a snow flake cause a flood?}}\\
Capable of:       &\multicolumn{8}{l}{{\it Can a helmet be driven?}}\\
Default inheritance:&\multicolumn{8}{l}{{\it Did Napoleon have a mouth?}}\\
Precondition:     &\multicolumn{8}{l}{{\it Is a sunrise possible only after seeing a cloud?}}\\
Boundary (bona fide):&\multicolumn{8}{l}{{\it Is a trouser waist belt joined to a leg opening?}}\\
Antonym:         &\multicolumn{8}{l}{{\it Does pursue have an opposite meaning to retreat?}}\\
Contained in:     &\multicolumn{8}{l}{{\it Can a credit card be contained in a wool fibre?}}\\
Has member:     &\multicolumn{8}{l}{{\it Is haiku a type of art?}}\\
Troponym:       &\multicolumn{8}{l}{{\it Is crushing a type of breaking?}}\\
Boundary (fiat):&\multicolumn{8}{l}{{\it Is Iran joined to Pakistan?}}\\
\Xhline{4\arrayrulewidth}
    \end{tabular}
    }
    \vspace{-3mm}
\end{table}

\section{Analogical Reasoning Tests A \& B}

Drawing on objective reality we craft a high-quality physical common sense benchmark that is balanced and stratified for its relation types with verifiable common knowledge. 

\subsubsection{Two Tasks.} 
Due to its importance in everyday reasoning we expect analogies to be one of the central mechanism for applying core physical common sense to new situations \cite{sternberg1977component,hofstadter2013surfaces}. Two  experimental tasks were employed to establish norms: inductive reasoning with uncued cross-domain analogies (Section A: 48 unique questions), and analogy relations (Section B: 320 unique questions) that underlie the analogies. The questions were designed using expert introspection, collectively agreed after independent answering by the co-authors, and Google checking. Both ART A and B had an equal balance of the 16 semantic relations (described in Table~\ref{tab:datasets}, lower part) that we discuss below. Overall, approximately half the questions in each analogy and each analogy relation were judged by the paper authors to be Yes (True) and half were judged to be No (False).\footnote{ART A: 25 Yes, 23 No with 8 joint resolutions; ART B: 154 Yes, 166 No with 52 joint resolutions.} A simple and consistent syntax was chosen for all items within a semantic relation. 

Consider this example of an analogy question: {\it Do congeal and thicken have a similar relationship to whisper and murmur?} The expected answer would be Yes (True). Analogical reasoning \cite{sternberg1977component,gentner2001analogical} requires participants to grasp that the source entities (e.g. \lq congeal' and \lq thicken') are related by a specific relationship (e.g. synonymy) based on their shared attributes. Participants then need to map this source analog to a target domain. 
In the case of \lq False' questions, the second pair of words would be related by a different semantic relation from within the set of 16 types. 
The following is an example of an analogy relation question: {\it Is coal shinier than aluminium foil?}  The expected answer would be No (False). In terms of intensity, \lq coal' is less shiny than \lq aluminium foil'. For analogy relation questions, questions judged True and False by ourselves were considered to represent expert truth and falsehood about objective reality, respectively. 

\subsubsection{Semantic Relations.}\label{sec:semanticrelations}
We now turn to consider an inventory of physical common sense relationships that we expect to be present within the core of common sense reasoning. 

The strength and stability of norms within this inventory will be tested by our human subject study. As reported in Table~\ref{tab:datasets}~(lower part), the semantic relations are all general purpose, can be applied across domains, and various subsets have been widely used in prior classification schemes \cite{bejar2012cognitive,spradley2016participant}. We leave to future work, the intra-rater stability of the scores over time, as well as a test over a wider demographic. 
Each of the 16 relations is measured by 20 questions which require a Yes/No answer. The choice of questions is based on an intuition that each of the semantic relations contributes to an understanding of the physical world. Each question was broadly intended to surprise readers so that answering would engage both knowledge and processes beyond association and rote learning. For example, comparing two objects for size at vastly different scales or in highly unlikely domains, or considering whether possible attributes such as colour were necessary attributes of an object. Questions were designed also to contain misleading cues, for example asking about whether sibling concepts are part of each other. In this respect the questions try to create a mildly hostile environment although not one that is actively informed by adversarial testing on any model. Note that analogies were not restricted by domain, e.g. {\it Do jeans and cotton have a similar relationship to tree and leaf?}

\subsubsection{Web Checking.}\label{sec:gcheck}
We eliminated candidates with exact or trivial variations of the question and answer on the Web, arriving at the final set of 48 ART A and 320 ART B questions. 

\subsubsection{Equivalence and Gold Standard.} To obtain an objective benchmark, a panel of three experts (co-authors) independently annotated
the tests with a Fleiss's $\kappa$ of 0.75 (8 item disagreements giving moderate-to-substantial agreement with 95\% CI) on Section A and 0.77 (52 item disagreements giving substantial agreement with 95\% CI) on Section B.  Sources of difference were resolved by discussion to produce a gold standard by making minor modifications to the test question or by agreeing a single interpretation. All disagreements could be resolved. Disagreements were due to word sense understanding, e.g. {\it impure} as mixed with foreign matter vs. morally corrupt, semantic errors in the question, e.g. a comparison to {\it deck} vs. {\it keel} where spatial orientation was important, and the exact degree of probability implied by modal verbs such as {\it will X cause Y}, ranging from tentative to required. 
\section{Methods}

\noindent{\bf Participants.} 63 participants were recruited from Prolific Academic 
with at least a master degree who were native English speakers resident in the UK.\footnote{Ethical approval was granted by  University of Cambridge's Modern and Medieval Languages and Linguistics  Ethics committee on condition of informed consent, anonymisation, and secured storage.} Two participants were excluded due to non-completion. Hence 61 participants, aged 22 to 68 (mean = 34.4, SD = 10.4) where 13 identified their gender as male and 48 as female, were our subject group. The participants were given all questions to complete in one hour and ten minutes, with a recommendation to take a rest every 40 to 50 questions. Subjects were free to go over the guide time up to a limit of three hours.   Questions were presented in random order. For each test we provided four complete and three warm up examples. In practice all subjects exceeded 70\% without being asked to re-do the test. The participants were asked to not use other sources such as the Web.

\noindent{\bf LLMs.} 
We used five variants of GPT-3 with different parameter sizes, Ada (350M), Babbage (1.3B), Curie (6.7B), text-Davinci-002 (175B), code-Davinci-002 (175B) and ChatGPT (175B). For all GPT-3 models, to ensure all models' performance reflects their parameter sizes, and not their ability to follow the instructions,
we used the conditional probability of ``Yes'' and ``No'' tokens given a prompt to select model's response. The temperature parameter for models was set to 0 in all experiments, switching off stochastic behavior. The prompts are either Direct questions, e.g. \emph{Does a missing wheel stop a car from running?} or Categorical questions including the semantic relations, e.g. \emph{In terms of order of size, is a country longer than a town?} For the code-Davinci-002 and ChatGPT models, we also experimented with Chain-of-Thoughts~\cite{kojima2022large,wei2022chain} prompting (denoted as CoT). Default prompting style is asking Direct questions unless stated otherwise.

\section{Results}

\subsection{Human Variation on the ART A\&B\footnote{We expect human variation due to performance errors \cite{chomsky2006language} or algorithmic differences \cite{cherniak1990minimal}.}}

\noindent{\bf Ceiling/Floor Effect.} We adopted the common threshold to test for ceiling and floor effects. These were considered present if 15\% of the human subjects achieved the worst (chance level) or the best score on ART A or B as a whole. A secondary objective was to test for ceiling/floor effects on any subset of ART B. No ceiling or floor effect was observed for either of the tests as a whole. However a floor effect was observed for {\it Has part} and a ceiling effect was observed for {\it Order of intensity}, {\it Order of size} and {\it Cause and effect}.

\noindent{\bf Internal Consistency.} Since our test items are dichotomous with varying levels of difficulty, reliability was measured using the Kuder-Richardson Formula 20 ({\it KR}$_{20}$) on the collectively agreed gold answers. The value for ART A for all questions was 0.72 (moderate/high), and for ART B was 0.87 (high) indicating the adequacy of the tests.

\noindent{\bf Establishing Levels of Agreement.} Table \ref{tab:res1} shows human subject performance against the gold standard. The 61 participants scored a mean of 36.3 correct answers (76\%) on ART A and 260.5 (81\%) on ART B. We further examined our data and noted cases where a large majority ($\ge80\%$) disagreed with the gold standard (2 on ART A and 7 on ART B). 

\noindent{\bf Relationship between Part A and B.} We used the Jarque-Bera non-normality test taking account of both skewness and kurtosis to determine with $p<0.05$ that human subject scores on both ART A\&B were not normally distributed. On inspection both had heavy left tails (Skewness: ART A -0.89, ART B -1.32) and ART B was leptokurtik (Kurtosis: ART A 0.25, ART B 3.60). We therefore employed Spearman rank correlation coefficient and found moderate evidence ($\rho=0.46$) for dependence between the rankings in performance.

\begin{table}[t!]
    \centering\small
        \caption{Human performance on the ART  A\&B. AG@N\% is the percentage of questions on which a $N\%$ of participants agreed. Maximum agreement is 48 for the part A, 320 for part B, and 20 for each semantic relation. $^{1}$ shows the range of correct answers that are possible for individual participants where the gold standard is the expert committee; $^{2}$ shows the range of correct answers for individual participants where the gold standard is the expert committee; $^{3,4}$ Show the mean and standard deviation over all participants; $^{5}$ C/F@15 reports the ceiling or floor effect, which is reported as No if less than 15\% of test takers (9 out of 61) obtained the maximum or minimum possible scores, otherwise Yes; $^{6}$ A composite score across all the semantic relations.}
    \label{tab:res1}
    
    \scalebox{0.67}{
\begin{tabular}{l p{0.8cm} p{0.9cm} p{1.2cm} p{0.5cm} p{0.4cm} p{0.7cm} p{0.6cm} p{0.6cm}}
\toprule
  & Missing & Potential & Observed & M$^{3}$ & SD$^{4}$  
  & C/F$^{5}$  & AG & AG\\ 
Measure  & \% & range$^{1}$ & range$^{2}$ & & %&  
& @15\%  & @80\%&@90\%  \\

\midrule
Analogy Test & 0\% & 0--48 & 22--43 & 36.3 & 4.6 %& 0.697 
& N/N & 26 & 14 \\
\midrule
Analogy Relation Test$^{6}$ & 0\% & 0--320 & 199--285 & 260.5 & 15.5 %& 0.870  
& N/N & 228 & 183 \\
\midrule
1.Synonym & 0\% & 0--20 & 11--20 & 16.4 & 2.0 %& 0.509 
& N/N & 14 & 11 \\
2.Necessary quality  & 0\% & 0--20 & 12--19 & 15.7 & 1.7 %& 0.268 
& N/N & 12 & 10 \\
3.Associated quality &  0\% & 0--20 &10--18 & 14.8 & 1.6 %& 0.135 
& N/N & 15 & 10 \\
4.Has part &  0\% & 0--20 &  10--19 & 15.5 & 1.6 %& 0.365 
& Y/N & 15 & 12 \\
5.Order of intensity & 0\% & 0--20 & 12--20 & 18.7 & 1.5 %& 0.531 
& N/Y & 19 & 16 \\
6.Order of size & 0\% &0--20 & 14--20 & 18.8 & 1.4 %& 0.473 
& N/Y & 19 & 18 \\
7.Cause and effect & 0\% & 0--20 & 13--20 & 18.0 & 1.6 %& 0.420 
& N/Y & 16 & 13 \\
8.Capable of & 0\% & 0--20 &  12--20 & 16.2 & 1.8 %& 0.452 
& N/N & 12 & 12 \\
9.Default inheritance & 0\% &0--20 & 12--19 & 16.7 & 1.5 %& 0.303 
& N/N & 16 & 14 \\
10.Precondition & 0\% &0--20 & 11--20 & 17.0 & 1.7 %& 0.369 
& N/N & 15 & 11 \\
11.Boundary (bona fide) & 0\% &0--20 &  9--19 & 14.6 & 2.1 %& 0.363 
& N/N & 13 & 6 \\
12.Antonym &  0\% &0--20 & 9--18 & 15.2 & 2.1 %& 0.358 
& N/N & 12 & 6 \\
13.Contained in & 0\% & 0--20 & 14--20 & 17.6 & 1.4 %& 0.042 
& N/N & 15 & 14 \\
14.Has member & 0\% & 0--20 &10--19 & 15.2 & 2.0 %& 0.494 
& N/N & 12 & 11 \\
15.Troponym & 0\% & 0--20 &14--20 & 17.4 & 1.4 %& 0.169 
& N/N & 16 & 14 \\
16.Boundary (fiat) & 0\% & 0--20 &7--17 & 13.0 & 2.1 %& 0.242 
& N/N & 7 & 5  \\
\bottomrule
    \end{tabular}
    }
    \vspace{-3mm}
\end{table}

\subsection{Identifying a Common Sense Subset of ART A\&B} 

We cannot simply assume that different people perceive all the relationships between concepts in our test set in the same way. We therefore try to establish the common sense norms of na\"ive physics by establishing the set of questions on which a majority of participants agreed. In the common sense subset, the gold answers are the majority answers chosen by the human subjects. Table \ref{tab:res1} shows the number of questions on which participants agreed, both as a composite and grouped by individual semantic relations. Agreement@80\%, the value we choose for our common sense subset, shows the number of questions on which 80\% or more of participants agreed. After filtering for 80\% agreement threshold, 26 out of 48 ART A questions (54\%) remained and 228 out of 320 ART B questions (71\%) remained for evaluating LLMs.

\subsection{LLMs vs. the Full Set of ART A\&B}

Having established an approximation of a human norm we compare LLMs to human performance. Since LLMs do not exhibit random performance errors due to the environment, we consider that any difference between the LLM and the human subject reference interval exhibits a potential divergence in rationality. Table \ref{tab:res3} highlights both alignment and divergence on na\"ive physical common sense understanding between LLMs and human subjects on ART A\&B.

\noindent{\bf LLMs on ART A.} Babbage, code-Davinci-002 and ChatGPT had composite scores that were within the lower part of the reference range, whilst Ada, Curie and text-Davinci-002 fell outside the reference range. Performing chain-of-thought (CoT) prompting with categories (Categorical) increased performance, bringing small improvements to code-Davinci-002 and ChatGPT.%
 Despite the difference in question formulation, these findings are consistent with 
 what was observed in the zero-shot SAT Analogy experiments of
 \cite{brown2020language}.

\noindent{\bf LLMs on ART B.} Ada, Babbage, Currie and text-Davinci-002 composite scores for ART B fell under the reference intervals. For text-Davinci-002 the composite score is approaching human levels but this result obscures an uneven understanding across categories. For text-Davinci-002, {\it Order of intensity}, {\it Order of size}, {\it Precondition}, {\it Necessary quality}, and {\it Fiat boundary} show marked divergence to the reference intervals. This outcome supports the view of Aroca-Ouellette {\it et al.} \shortcite{aroca2021prost} on GPT-1 and -2 that such models are challenged on object attributes (e.g. height) and affordances (e.g. breakability). Moving to ChatGPT we observed that performance has improved markedly for {\it Necessary quality}, {\it Associated quality}, {\it Order of intensity} and {\it Precondition}. We speculate that this is a result of the combination of careful selection of training data (e.g. instructional texts) as well as reinforcement learning. A group of ChatGPT tests showed quantitative performance at the upper end of the reference interval on {\it Necessary quality}, {\it Associated quality}, {\it Precondition}, {\it Troponym} and somewhat surprisingly {\it Cause and effect}. This result provides a challenge to the view that only rationalistic (i.e. logic based) language models can be successful in answering common sense questions correctly. Yet we should be cautious in accepting this result at its face value. As we discuss later, models draw on a wide range of associations that sometimes allows them to display seemingly correct reasoning. 
Such effects however are hard to isolate.

\begin{table*}[t!]
    \centering\small
        \caption{LLM performance on the full set and common sense subet of ART A\&B where green shows performance within the reference range and pink outside the reference range. $^{1}$ Shows the potential range of correct answers that are possible for individual participants where the gold standard is the expert committee; $^{2}$ Shows the reference interval calculated as the range within which 95\% of the human subject reference population values fall. $^+$ Later we run another test on the sensitivity of the results on the order of arguments in nonsymmetric relations by reversing the order of arguments, and observe the inability of the model to correctly answer \emph{Cause and Effect} and \emph{Troponym} questions.}
    \label{tab:res3}
        \scalebox{0.58}{
\begin{tabular}{p{2.7cm} p{1cm} p{1.2cm} p{0.3cm} p{0.9cm} p{0.7cm} p{1.5cm} p{0.7cm}  p{1.3cm}p{0.7cm}p{1.2cm}p{1cm} c p{0.3cm} p{0.9cm} p{0.7cm} p{1.5cm} p{0.7cm}  p{1.3cm}p{0.7cm}p{1cm}}
\toprule
&  \multicolumn{10}{c}{\bf{Full ART A\&B}} &  \multicolumn{10}{c}{\bf{Common Sense Subset of ART A\&B}}  \\
           \cmidrule(lr){2-11}\cmidrule(lr){12-21}
           & \multicolumn{2}{c}{\bf{Range}} & \multicolumn{4}{c}{\bf{GPT-3}}&\multicolumn{2}{c}{\bf{code-davinci+CoT}}&\multicolumn{2}{c}{\bf{ChatGPT}}   & \multicolumn{2}{c}{\bf{Range}} & \multicolumn{4}{c}{\bf{GPT-3}}&\multicolumn{2}{c}{\bf{code-davinci+CoT}}&\multicolumn{2}{c}{\bf{ChatGPT}} \\
           \cmidrule(lr){2-3} \cmidrule(lr){4-7}\cmidrule(lr){8-9}\cmidrule(lr){10-11} \cmidrule(lr){12-13}\cmidrule(lr){14-17}\cmidrule(lr){18-19}\cmidrule(lr){20-21}
  Measure  & Potential$^{1}$   & Reference$^{2}$   & Ada    & Babbage & Currie   & text-davinci & Direct & Categorical  & Direct& Categorical  &Potential$^{1}$   & Reference$^{2}$   & Ada    & Babbage & Currie   &text-davinci&  Direct & Categorical &Direct&Categorical\\
\midrule
ART Section A & 0--48 &  28--43 & \cellcolor{pink}22 & \cellcolor{lime}32 & \cellcolor{pink}24 & \cellcolor{pink}27& \cellcolor{lime}28 & \cellcolor{lime}32 &\cellcolor{lime}30&\cellcolor{lime}32&0--26 & 18--25 & \cellcolor{pink}9 & \cellcolor{lime}21 & \cellcolor{lime}19 & \cellcolor{pink}15  & \cellcolor{pink}17 & \cellcolor{pink}16 &\cellcolor{lime}18&\cellcolor{lime}19\\
\midrule
ART Section B & 0--320 & 231--285  & \cellcolor{pink}168 & \cellcolor{pink}169  & \cellcolor{pink}177 & \cellcolor{pink}221 & \cellcolor{lime}238&\cellcolor{lime}241 & \cellcolor{lime}250&\cellcolor{lime}255&0--228 & 192--220 & \cellcolor{pink}148& \cellcolor{pink}135 & \cellcolor{pink}137 & \cellcolor{pink}180 & \cellcolor{pink}191 & \cellcolor{pink}191 &\cellcolor{lime}203&\cellcolor{lime}202\\
\midrule
Synonym & 0--20 &  13--20 & \cellcolor{pink}11 &  \cellcolor{pink}10 & \cellcolor{pink}12 & \cellcolor{lime}17& \cellcolor{lime}16 &\cellcolor{lime}17 &\cellcolor{lime}17&\cellcolor{lime}17& 0--14 & 12--14 & \cellcolor{pink}10 & \cellcolor{pink}10 & \cellcolor{pink}8 & \cellcolor{lime}12  & \cellcolor{lime}14 & \cellcolor{lime}14 &\cellcolor{lime}13&\cellcolor{lime}13\\
Necessary quality & 0--20 & 13--18 & \cellcolor{pink}12 & \cellcolor{lime}13 & \cellcolor{lime}13 & \cellcolor{pink}11& \cellcolor{lime}13 & \cellcolor{lime}14 &\cellcolor{lime}18&\cellcolor{lime}19&0--12 &11--12 & \cellcolor{pink}9 & \cellcolor{pink}7 & \cellcolor{pink}8 & \cellcolor{pink}9  & \cellcolor{pink}10 &\cellcolor{pink}10 &\cellcolor{lime}11&\cellcolor{lime}11\\
Associated quality &  0--20 & 12--17 & \cellcolor{pink}8 & \cellcolor{pink}7 & \cellcolor{pink}8 & \cellcolor{lime}15& \cellcolor{lime}13 & \cellcolor{lime}12 &\cellcolor{lime}14&\cellcolor{lime}16 & 0--15 & 10--14 & \cellcolor{lime}10 & \cellcolor{pink}6 & \cellcolor{pink}4 & \cellcolor{lime}14& \cellcolor{lime}11 & \cellcolor{lime}10 &\cellcolor{lime}13&\cellcolor{lime}13\\
Has part & 0--20 & 13--18 & \cellcolor{pink}11 & \cellcolor{pink}11 & \cellcolor{pink}11 & \cellcolor{lime}13& \cellcolor{lime}15 & \cellcolor{pink}12 &\cellcolor{lime}14& \cellcolor{lime}14&  0--15 &12--15 & \cellcolor{pink}11 & \cellcolor{lime}12 & \cellcolor{lime}12 & \cellcolor{lime}13 & \cellcolor{lime}13 & \cellcolor{lime}12 &\cellcolor{lime}13&\cellcolor{lime}13\\
Order of intensity & 0--20 &16--20 & \cellcolor{pink}10 & \cellcolor{pink}12 & \cellcolor{pink}10 & \cellcolor{pink}12& \cellcolor{lime}16 & \cellcolor{pink}15 &\cellcolor{lime}19&\cellcolor{lime}17& 0--19 & 16--19 & \cellcolor{pink}10 & \cellcolor{pink}11 & \cellcolor{pink}9 & \cellcolor{pink}11  & \cellcolor{pink}15 & \cellcolor{pink}14 &\cellcolor{lime}18&\cellcolor{lime}16\\
Order of size & 0--20 &  17--20 & \cellcolor{pink}10 & \cellcolor{pink}9 & \cellcolor{pink}10 & \cellcolor{pink}11& \cellcolor{pink}13 &\cellcolor{pink}14 &\cellcolor{pink}13&\cellcolor{pink}16& 0--19 & 16--19 & \cellcolor{pink}10 & \cellcolor{pink}9 & \cellcolor{pink}10 & \cellcolor{pink}11  & \cellcolor{pink}13 & \cellcolor{pink}14 &\cellcolor{pink}13&\cellcolor{lime}16\\
Cause and effect & 0--20 &  15--20 & \cellcolor{pink}9 & \cellcolor{pink}12 &  \cellcolor{pink}10 & \cellcolor{lime}$18^+$& \cellcolor{lime}$18^+$ & \cellcolor{lime}$18^+$ &\cellcolor{lime}$19^+$&\cellcolor{lime}$18^+$ & 0--16 &14--16 & \cellcolor{pink}8 & \cellcolor{pink}10 & \cellcolor{pink}10 & \cellcolor{lime}15 & \cellcolor{lime}15 & \cellcolor{lime}15 &\cellcolor{lime}16&\cellcolor{lime}15\\
Capable of & 0--20 &  13--19 & \cellcolor{pink}12 & \cellcolor{pink}11 & \cellcolor{pink}9 & \cellcolor{lime}17& \cellcolor{lime}14 & \cellcolor{lime}13 &\cellcolor{lime}15&\cellcolor{lime}15 & 0--12 & 11--12 & \cellcolor{pink}10 & \cellcolor{pink}3 & \cellcolor{pink}4 & \cellcolor{lime}11  & \cellcolor{lime}11 & \cellcolor{lime}11 &\cellcolor{lime}11&\cellcolor{lime}11\\
Default inheritance & 0--20 & 14--19 & \cellcolor{pink}10 & \cellcolor{pink}11 & \cellcolor{pink}13 & \cellcolor{lime}17& \cellcolor{lime}16&\cellcolor{lime}17 &\cellcolor{lime}18&\cellcolor{lime}17&  0--16 & 13--16 & \cellcolor{pink}10 & \cellcolor{pink}11 & \cellcolor{lime}13 & \cellcolor{lime}16 & \cellcolor{lime}14 & \cellcolor{lime}15 &\cellcolor{lime}15&\cellcolor{lime}14\\
Precondition & 0--20 &  14--20 & \cellcolor{pink}11 & \cellcolor{pink}8 & \cellcolor{pink}11 & \cellcolor{pink}10& \cellcolor{lime}18 &\cellcolor{lime}18 &\cellcolor{lime}16&\cellcolor{lime}19& 0--15 & 13--15 & \cellcolor{pink}11 & \cellcolor{pink}8 & \cellcolor{pink}11 & \cellcolor{pink}10 & \cellcolor{lime}14 & \cellcolor{lime}13 &\cellcolor{lime}13&\cellcolor{lime}14\\
Boundary (bona fide) & 0--20 & 11--17 & \cellcolor{lime}11  & \cellcolor{lime}14 & \cellcolor{lime}12 & \cellcolor{lime}12 & \cellcolor{lime}11 &\cellcolor{lime}16 &\cellcolor{lime}14&\cellcolor{lime}16& 0--13 & 9--12 & \cellcolor{pink}8 & \cellcolor{pink}8 & \cellcolor{pink}8 & \cellcolor{lime}10 & \cellcolor{lime}9 & \cellcolor{lime}11 &\cellcolor{lime}10&\cellcolor{lime}10\\
Antonym & 0--20 & 12--18 & \cellcolor{pink}10 &\cellcolor{pink} 5 & \cellcolor{pink}9 & \cellcolor{lime}14 & \cellcolor{lime}14 &\cellcolor{lime}14 &\cellcolor{lime}16&\cellcolor{lime}16&  0--12 & 8--12 & \cellcolor{pink}7 & \cellcolor{pink}3 & \cellcolor{pink}5 & \cellcolor{lime}9 & \cellcolor{lime}8 & \cellcolor{lime}9 &\cellcolor{lime}10&\cellcolor{lime}11\\
Contained in & 0--20 & 15--20 & \cellcolor{pink}10  & \cellcolor{pink}14 &  \cellcolor{pink}13 & \cellcolor{pink}13& \cellcolor{pink}14 &\cellcolor{pink}13 &\cellcolor{lime}15&\cellcolor{pink}14  & 0--15 & 13--15 & \cellcolor{pink}9 & \cellcolor{pink}12 & \cellcolor{pink}11 & \cellcolor{pink}10 & \cellcolor{pink}11 &\cellcolor{pink}10 &\cellcolor{lime}13&\cellcolor{pink}11\\
Has member & 0--20 & 12--19 & \cellcolor{pink}11 & \cellcolor{pink}11 & \cellcolor{lime}14 & \cellcolor{lime}16& \cellcolor{lime}17 & \cellcolor{lime}17 &\cellcolor{lime}15&\cellcolor{lime}14  & 0--12 & 10--12 & \cellcolor{lime}10 & \cellcolor{lime}11 & \cellcolor{lime}10 & \cellcolor{lime}12& \cellcolor{lime}11 & \cellcolor{lime}11 &\cellcolor{lime}11&\cellcolor{lime}11\\
Troponym & 0--20 & 15--20 & \cellcolor{pink}11 & \cellcolor{pink}10 & \cellcolor{pink}13 & \cellcolor{lime}$16^+$& \cellcolor{lime}$17 ^+$&\cellcolor{lime}$18^+$ &\cellcolor{lime}$17^+$&\cellcolor{lime}$18^+$ & 0--16 & 14--16 & \cellcolor{pink}11 & \cellcolor{pink}10 & \cellcolor{pink}12 & \cellcolor{pink}13 & \cellcolor{lime}16 & \cellcolor{lime}16 &\cellcolor{lime}16&\cellcolor{lime}16\\
Boundary (fiat) &  0--20 & 10--17 & \cellcolor{lime}11 & \cellcolor{lime}11 & \cellcolor{pink}9 & \cellcolor{pink}9& \cellcolor{lime}13 & \cellcolor{lime}13 &\cellcolor{lime}10&\cellcolor{pink}9&  0--7 & 5--6  & \cellcolor{pink}4 & \cellcolor{pink}4 & \cellcolor{pink}2 & \cellcolor{pink}4 & \cellcolor{lime}5 & \cellcolor{lime}5 &\cellcolor{lime}6&\cellcolor{lime}6\\
\bottomrule
    \end{tabular}
    }
    \vspace{-2mm}
\end{table*}

\subsection{LLMs vs. the Common Sense Subset of ART A\&B}

The Table~\ref{tab:res3}~(right block) shows the LLM performance on the common sense subset of ART A\&B. Reference ranges have unsurprisingly narrowed compared to the full tests, and unfortunately for two relations ({\it Antonym} and {\it Fiat Boundary}) the lower reference bound falls very close to the random baseline. As shown by the upper value of the potential range, several relations appear to have markedly strong consensus at $\ge80\%$ agreement among human subjects including {\it Order of intensity}, {\it Order of size}, {\it Cause and effect}, {\it Precondition}, {\it Contained in} and {\it Troponym}. On the other hand {\it Capable of}, {\it Bona fide boundary} and {\it Antonym} had fewer consensus items at 80\% and {\it Fiat boundary} had the fewest consensus items indicating how disordered the judgements were in this category. The common sense subset favoured ChatGPT over other models showing that a combination of reinforcement learning and data selection led to performance on a par with humans in many categories. 
The subset revealed churn on a category by category basis. Across all semantic relations Curie dropped out of three reference ranges and gained two, Babbage dropped out of three and gained two, Ada dropped two and gained two. 
Percentage improvements on ART B
reveal modest gains for all models on the common sense subset (Ada +12\%, Babbage +6\%, Curie +5\%, text-Davinci-002 +10\%, code-Davinci-002 +9\%, ChatGPT +11\%).

\section{Discussion}\label{sec:discussion}
Our investigation has drawn attention to several algorithmic deficiencies in the LLMs to reason about objects in the physical world. We highlight a few that warrant more  investigation: 

\noindent{\bf Mereo-Topology.} Humans understand objects at various granular levels.
Although we cannot be entirely sure of ChatGPT's algorithm, ART seems to demonstrate that GPT-3 using text alone without reinforcement learning from human judgements performs at the lower end or outside the reference interval for {\it Has part} and {\it Contained in}. We tend to discount the result for {\it Boundary (fiat)} due to the low level of agreement among human subjects. Interestingly, GPT-3 performance on demarcating boundaries appears less variant to model size, indicating that a combination of more data with ever larger numbers of model parameters is not a viable pathway to improvements.

\noindent{\bf Affordance Characteristics.} 

Object properties seem to be understood by humans in terms of their affordance characteristics, or in other words the opportunities and limitations that they offer to humans \cite{gibson1979ecological,smith2003mountains}. ART has shown up weaknesses in GPT-3 models to understand affordance characteristics through {\it Necessary quality}, {\it Order of size} and {\it Order of intensity}. Severe performance limitations on {\it Order of size} appear consistent with previous reports \cite{talmor2022commonsenseqa}, and supports the view that there is an algorithmic weakness in the ability of language-only distributed semantic representations to capture affordance characteristics \cite{glenberg2000symbol,jones2022distrubutional}. Our finding is particularly salient in the light of strong human performance and low human variance for {\it Order of size} and {\it Order of intensity}. ChatGPT's progress on {\it Necessary and Associated quality} shows how reinforcement learning has been able to skillfully leverage human judgements to compensate for weaknesses in the data.

In other areas of object understanding GPT-3 appears to offer evidence for verbal reasoning on a par with our human subjects. For example in {\it Synonym}, {\it Antonym}, {\it Default inheritance}, and intriguingly, {\it Cause and effect}. We take this finding not as a conclusion but as the next starting point for further investigation. 

\noindent{\bf Stability on Nonsymmetric Semantic Relations.} In order to test whether the responses arise from associative fortuity~\cite{pearl2009causality} in the data we tested Davinci and ChatGPT sensitivity to argument reversal for nonsymmetric relations (e.g. Original: {\it Does a missing wheel stop a car from running}; Modified: {\it Does a car stop a missing wheel from running?}) and seeing if this changed Davinci's response. Results of this on 9 semantic relations with nonsymmetric items (i.e., \emph{Cause and effect}, \emph{Order of intensity}, \emph{Order of size}, \emph{Has part}, \emph{Default inheritance}, \emph{Precondition}, \emph{Contained in}, \emph{Has member}, \emph{Troponym})  indicated that both Davinci and ChatGPT responses are stable under 7 relations but exhibit potential signs of associative behavior for \emph{Cause and effect} and \emph{Troponym} with GPT-3 and ChatGPT failing (e.g., replying yes for both of the examples provided above) on 70\% and 50\% of the cases, respectively. Combining LLMs with symbolic~\cite{nye2021improving} or physical world~\cite{ahn2022can} grounding are promising directions to address these side effects of associative learning.

\section{Conclusion}

This work builds on foundations that are already firmly established in different fields: the importance of lexical semantic relationships for language processing \cite{miller1995wordnet}, the importance of conceptual modeling based on basic ontological principles \cite{smith2012ontology}, and prior evidence that establishes the characteristics and norms of human reasoning \cite{stanovich1998individual}. 

We have questioned therefore the strong empirical assumption of much prior work in NLP about the ability of LLMs to represent physical common sense using language data alone \cite{lake2017building} and identified some promising areas for algorithmic investigation and theory formation. Going forward we hope that the discussion in this paper will be a useful adjunct for intrinsic evaluation of natural language generation models and further to LLMs that aim to combine machine learning from text with nature-inspired cognitive capabilities.

\bibliographystyle{apacite}

\setlength{\bibleftmargin}{.125in}
\setlength{\bibindent}{-\bibleftmargin}

\bibliography{CogSci_Template}

\end{document}